\documentclass{article}

\PassOptionsToPackage{numbers, compress}{natbib}

\usepackage[preprint]{neurips_2025}

\usepackage[utf8]{inputenc} 
\usepackage[T1]{fontenc}    
\usepackage{hyperref}       
\usepackage{url}            
\usepackage{booktabs}       
\usepackage{amsfonts}       
\usepackage{nicefrac}       
\usepackage{microtype}      
\usepackage{xcolor}         
\usepackage{tikz}
\usetikzlibrary{arrows.meta} 
\usepackage{algorithm,algpseudocode}

\usepackage{mathrsfs,mathtools}
\usepackage{amsmath,amssymb,amsthm,enumitem,amsfonts,nccmath}
\usepackage{cleveref}
\usepackage{subcaption}
\usepackage{fontawesome}
\usepackage{cleveref}

\crefname{equation}{Equation}{Equations}
\crefformat{equation}{#2Equations~#1#3}
\crefrangeformat{equation}{#3Equations~#1#4 to~#5#2#6}
\crefmultiformat{equation}{#2Equations~#1#3}{ and~#2#1#3}{, #2#1#3}{ and~#2#1#3}
\newenvironment{tightemize}
{\begin{itemize}\itemsep1pt \parskip0pt \parsep0pt}{\end{itemize}\vspace{-\topsep}}

\usepackage{amsmath,amsfonts,bm}

\def\eqref#1{equation~\ref{#1}}

\def\1{\bm{1}}

\DeclareMathAlphabet{\mathsfit}{\encodingdefault}{\sfdefault}{m}{sl}
\SetMathAlphabet{\mathsfit}{bold}{\encodingdefault}{\sfdefault}{bx}{n}

\usepackage{booktabs}
\usepackage{multirow}
\usepackage{graphicx}
\usepackage{caption}
\usepackage{wrapfig}

\title{Efficient semantic uncertainty quantification in language models via diversity-steered sampling}

\author{%
  Ji~Won Park \\
  Prescient Design, Genentech \\
  \texttt{park.ji\_won@gene.com} \\
  \And
  Kyunghyun Cho \\
  Prescient Design, Genentech \\
  Center for Data Science, New York University\\
  \texttt{cho.kyunghyun@gene.com} \\
}

\begin{document}

\maketitle

\begin{abstract}
Accurately estimating \emph{semantic} aleatoric and epistemic uncertainties in large language models (LLMs) is particularly challenging in free-form question answering (QA), where obtaining stable estimates often requires many expensive generations. We introduce a \textbf{diversity-steered sampler} that discourages semantically redundant outputs during decoding, covers both autoregressive and masked diffusion paradigms, and yields substantial sample-efficiency gains. The key idea is to inject a continuous semantic-similarity penalty into the model’s proposal distribution using a natural language inference (NLI) model lightly finetuned on partial prefixes or intermediate diffusion states. We debias downstream uncertainty estimates with importance reweighting and shrink their variance with control variates. Across four QA benchmarks, our method matches or surpasses baselines while covering more semantic clusters with the same number of samples. Being modular and requiring no gradient access to the base LLM, the framework promises to serve as a drop-in enhancement for uncertainty estimation in risk-sensitive model deployments.
\end{abstract}

\section{Introduction} \label{sec:introduction}
Large language models (LLMs) excel at generating fluent text yet remain prone to both intrinsic \textit{aleatoric} ambiguity and \textit{epistemic} gaps in their learned knowledge. The latter can lead to hallucinations---confident outputs that are factually incorrect. Quantifying these uncertainties is critical for building safe AI systems deployable in high-stakes applications. In free-form natural language generation (NLG) tasks like question answering, this is especially challenging, as lexically distinct responses can still be \textit{semantically equivalent}.

Estimating uncertainty in language generation often relies on drawing large IID sample sets, which often contain semantically redundant outputs and waste compute. For example, semantic entropy has been proposed to quantify \textit{aleatoric} uncertainty by clustering generated outputs into semantic equivalence classes \citep{kuhn2023semantic}, while mutual information computed via iterative prompting has been used to lower-bound certain forms of \emph{epistemic} uncertainty \citep{yadkori2024believe}. Despite their conceptual appeal, both approaches require extensive sampling to produce stable estimates, limiting their use in low-resource settings. Diversity-oriented heuristics such as temperature scaling or nucleus sampling \citep{holtzman2019curious}, on the other hand, do not account for semantics. More recently, \citet{aichberger2025improving} proposed a method that steers generation toward semantic diversity, though it relies on token substitutions and remains restricted to autoregressive models (ARMs). We aim at extending this line of work.

While uncertainty estimation methods have focused on ARMs, masked diffusion models (MDMs) have recently emerged as strong alternatives. These models extend masked language modeling by learning iterative denoising schedules to progressively resolve masked spans \citep{austin2021structured, lou2023discrete, shi2024simplified, sahoo2024simple, ou2024your, nie2024llada}. Despite achieving text quality on par with state-of-the-art ARMs, MDMs remain largely overlooked in the context of uncertainty quantification. 

We propose a unified, model-agnostic framework that (1) actively steers decoding away from semantically redundant hypotheses, (2) corrects the induced sampling bias via importance weighting, and (3) reduces estimation variance with control variates. Crucially, our sampler operates in both ARM and MDM settings using entailment-based penalties computed on partial continuations or masked spans. A single natural language inference (NLI) model, fine-tuned minimally with a new \texttt{[TRUNC]} token for prefixes or a \texttt{[MASK]} token for diffusion masks, enables live semantic scoring without altering the base LLM. Our experiments evaluate the method’s ability to quantify established proxies for aleatoric and epistemic uncertainties, demonstrating improved uncertainty estimation across diverse NLP tasks. Practical enhancements, including adaptive tuning of the diversity hyperparameter and online stopping based on estimator stability, further improve sample efficiency.

\begin{figure}[t]
    \centering
    \includegraphics[width=1.0\linewidth]{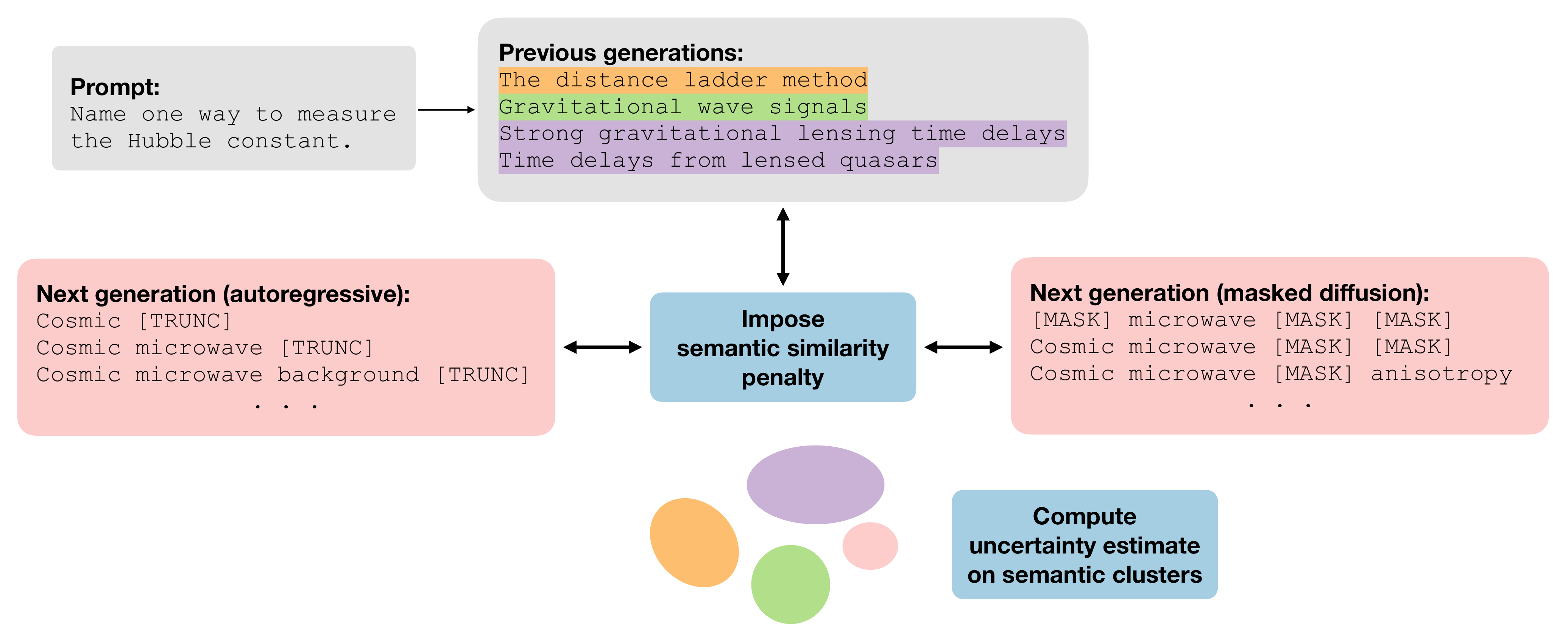}
    \caption{Our sampling workflow illustrated with a prompt that induces high aleatoric uncertainty. Given previous generations, we guide the LLM's next token away from semantically redundant outputs. The semantic clusters of resulting generations are used to estimate downstream uncertainty.} 
    \label{fig:overview}
\end{figure}

\section{Estimating the predictive uncertainty of free-form responses}

Let $x$ be an input (\textit{e.g.}, a question) and $\theta$ the weights of a pretrained language model. We wish to measure the total predictive uncertainty of the model’s output under semantic clustering. Following \citet{aichberger2025improving}, we define the distribution over semantic clusters $c\in\mathcal C$ by marginalizing over all output sequences $y$:
\begin{equation*}
p(c \mid x, \theta) =
\int 1[y \in c \mid x, \theta] \ p(y \mid x, \theta) dy,
\end{equation*}
which can be approximated with the MC estimator \footnote{The original presentation \citep{kuhn2023semantic} used the biased estimator, $p(c\mid x, \theta) = \sum_{y \in c} p(y \mid x, \theta)$.},
\begin{equation} \label{eq:cluster_prob_mc}
{\hat p}(c\mid x, \theta) \approx \frac{1}{N} \sum_{i=1}^N 1[y^{(i)} \in c], \quad y^{(1)}, \dotsc, y^{(N)} \sim p(y|x, \theta).
\end{equation}
If $y$ is a sequence of tokens $(y_1, \dotsc, y_T)$, then $p(y\mid x, \theta)$ can be computed as the product of individual conditional token probabilities, or, in terms of log probability: $\log p(y \mid x, \theta) =  \sum_{t=1}^N \log p(y_t \mid y_{<t}, x, \theta)$. 
The \emph{total} predictive uncertainty can then be written as the expected cross‐entropy between two independent draws of the model’s distribution \citep{aichberger2025improving}:
\begin{equation}
\label{eq:total_uncertainty}
\underbrace{\mathbb{E}_{\tilde \theta}\bigl[\mathrm{CE}\bigl(p(c\!\mid\!x, \theta)\,;\,p(c\!\mid\!x,\tilde \theta)\bigr)\bigr]}_{\rm total}
\;=\;
\underbrace{H\bigl(p(c\!\mid\!x, \theta)\bigr)}_{\rm aleatoric}
\;+\;
\underbrace{\mathbb{E}_{\tilde \theta}\Bigl[\mathrm{KL}\bigl(p(c\!\mid\!x, \theta)\,\big\|\,p(c\!\mid\!x,\tilde \theta)\bigr)\Bigr]}_{\rm epistemic},
\end{equation}
where $CE(\cdot;\cdot)$ is the cross entropy, $H(\cdot)$ is the shannon entropy, and $\mathbb{E}_{\tilde \theta} \coloneqq \mathbb{E}_{\tilde \theta \sim p(\cdot|D)}$ denotes expectation over the weight posterior.
As observed by \citet{aichberger2025improving}, the first term on the right,
\begin{equation} \label{eq:semantic_entropy}
H\bigl(p(c\mid x, \theta)\bigr)
\;=\;-\,\sum_{c\in\mathcal C}p(c\mid x, \theta)\,\log p(c\mid x, \theta),
\end{equation}
is the \emph{aleatoric semantic uncertainty}, also called semantic entropy (SE) \citep{kuhn2023semantic}. It captures the irreducible ambiguity in the meaning of valid outputs under a single model. The second term,
\begin{equation} \label{eq:semantic_epistemic_bayesian}
\mathbb{E}_{\tilde \theta}\Bigl[\mathrm{KL}\bigl(p(c\mid x, \theta)\,\big\|\,p(c\mid x,\tilde \theta)\bigr)\Bigr],
\end{equation}
is the \emph{epistemic semantic uncertainty}, measuring our ignorance about which cluster is correct due to lack of model knowledge or data coverage. 

Let us consider possible answers to the question, ``Name a way to measure the Hubble constant.'' Aleatoric uncertainty arises when multiple answers are all valid. Even a reliable model may output multiple \emph{correct} variations that reflect different measurement methods, for example:
    \begin{tightemize}
      \item ``The distance ladder method, which uses Cepheid variable stars and Type Ia supernovae as standard candles to determine distances and redshifts of galaxies.''
      \item ``Gravitational lensing time delays, where the time differences in light arrival from multiple images of a lensed quasar are used to infer cosmic distances and expansion rate.''
    \end{tightemize}
These two answers belong to distinct semantic clusters only to the extent that they report different measurement methods; under a single model they contribute to SE (aleatoric uncertainty) but do not indicate that the model lacks knowledge of the phenomenon. To reduce aleatoric uncertainty, the user may rephrase the question to remove ambiguity (\textit{e.g.}, ``Name a way to measure the Hubble constant using quasars.''). On the other hand, if the model truly lacks knowledge or is out‐of‐distribution, it may produce answers that conflict with scientific facts or admit ignorance, for example:
\begin{tightemize}
  \item ``Sure! One method to measure the Hubble constant is by analyzing the oscillation patterns of intergalactic neutrino winds using quantum parallax interferometry.'' (hallucination)
  \item ``Tracking the color shift of moonlight reflected off distant asteroids.'' (hallucination) 
  \item ``I'm sorry, you might need to consult a scientific source or expert in cosmology for that.
\end{tightemize}
These responses fall into semantically distinct clusters that reflect gaps in the model’s knowledge (high epistemic uncertainty), signaling that the model is not trustworthy and should abstain.

\section{Methods}

\subsection{Diversity-steered sampling}

Our proposed sampling scheme modifies token-level conditional distributions to explicitly discourage semantically similar samples. Let us omit the conditioning on $x, \theta$ for notational clarity and let $p(y_t|y_{<t})$ represent the original language model distribution for the $t^{th}$ token given previously generated tokens $y_{<t}$. Our (unnormalized) sampling distribution is:
\begin{align} \label{eq:modified_dist_ar}
    \log \tilde q(y_t \ | \ y_{<t}) = \log p(y_t \ | \ y_{<t}) - \lambda \max_{s \in \mathcal{S}} E(y_{\leq t}, s)
\end{align}
where $\mathcal{S}$ is the set of previously sampled sequences, and $E(\cdot,\cdot)$ is a score quantifying the degree of semantic similarity between the inputs. In words, the tilting term repels the current sample away from the most similar existing generation. Aggregation schemes other than max, such as mean or median, may alternatively be used for softer guidance. Note also that we do not require gradients from the scoring function, and any scoring function that can handle partial sequences in either input argument would work. For full consistency with the NLI model used to semantically cluster the generations downstream, we finetune the same NLI model to accept partial sequences, as detailed in \autoref{sec:finetune_nli}. Concretely, we opt for the bidirectional entailment score:
\begin{align} \label{eq:bidirectional_entailment}
    E(y_{\leq t}, s) = \nicefrac{1}{2} \left( \texttt{entailment}(y_{\leq t}, s) + \texttt{entailment}(s, y_{\leq t}) \right),
\end{align}
where \texttt{entailment} is the entailment probability reported by an NLI model finetuned to handle partially generated sequences. The pseudocode for our sampling strategy is given in Algorithm~\ref{alg:diversity_sampling}.

\begin{algorithm}
\small
\caption{Diversity‐steered autoregressive sampling}\label{alg:diversity_sampling}
\begin{algorithmic}[1]
\Require Prompt $x$; base model $p(\cdot\mid x)$; bidirectional NLI scorer $E(\cdot,\!\cdot)$ from \autoref{eq:bidirectional_entailment}, trained with a special marker $\texttt{[TRUNC]}$ for incomplete text; diversity penalty $\lambda$; number of samples $N$; candidate tokens $\mathcal{V}$
\Ensure Set of semantically diverse generations $\mathcal{S}$
\State $\mathcal{S} \gets \varnothing$
\For{$i = 1$ \textbf{to} $N$} 
  \State $\mathrm{prefix} \gets \mathrm{tokenize}(x)$
  \State $\mathrm{token} \gets \texttt{<BOS>}$
  \While{$\mathrm{token} \neq \texttt{<EOS>}$}
    \For{$\mathrm{next}$ \textbf{in} $\mathcal{V}$} \Comment{Can alternatively consider the top-$k$ tokens only}
        \State $\ell(\mathrm{next}) \gets \log p \bigl(\mathrm{next}\mid \mathrm{prefix}\bigr)$
        \Comment{Base-model logits}
        \State $\hat s \gets \mathrm{decode}(\mathrm{prefix} \Vert \mathrm{next}) \Vert \texttt{[TRUNC]}$
        \Comment{Mark that $\hat s$ is unfinished}
        \State $\pi \gets \displaystyle\max_{s\in \mathcal{S}} \ E\bigl(\hat s, s\bigr)$
        \Comment{Similarity score with the most similar existing generation}
        \State $\ell'(\mathrm{next}) \gets \ell(\mathrm{next}) - \lambda \pi$
        \Comment{Repel toward semantic novelty}
        \EndFor
    \State $\mathrm{token}\sim\mathrm{Categorical} \bigl(\mathrm{softmax}(\ell')\bigr)$
    \State $\mathrm{prefix} \gets \mathrm{prefix} \Vert \mathrm{token}$
  \EndWhile
  \State $s^{(i)} \gets \mathrm{decode}(\mathrm{prefix})$
  \State $\mathcal{S} \leftarrow\mathcal{S} \cup \{s^{(i)}\}$
\EndFor
\State \Return $\mathcal{S}$
\end{algorithmic}
\end{algorithm}

\paragraph{Extension to masked diffusion models.}
In MDMs, decoding proceeds by iteratively refining a partially masked sequence $y^{(t)}$ through denoising steps. At each step $t$, a subset of masked positions is selected for infilling. Our diversity-steered strategy applies by modifying the denoising distribution $p(y^{(t)}|y^{(t+1)})$ to discourage infillings that are semantically similar to those from previous trajectories.

To compute similarity, we construct an intermediate input $z^{(t)}$ by substituting the current proposal $y^{(t)}$ into the masked positions of $y^{(t+1)}$, and then evaluate the unnormalized distribution:
\begin{align} \label{eq:modified_dist_mdm}
    \log \tilde q(y^{(t)} \ | \ y^{(t+1)}) = \log p(y^{(t)} \ | \ y^{(t+1)}) - \lambda \max_{s \in \mathcal{S}} E(z^{(t)}, s).
\end{align} The NLI model here is finetuned to handle masked or partially masked spans. This allows our method to promote semantic diversity across the entire denoising trajectory. Algorithm~\ref{alg:diversity_sampling_mdm} provides the analogous pseudocode for MDMs.

\paragraph{Adaptively tuning the diversity parameter.} 
The diversity strength parameter $\lambda$ significantly influences the semantic novelty of generated samples: too small a value yields redundancy as in vanilla IID sampling, while too large risks unnatural, low-likelihood outputs. To navigate this trade-off, we adaptively tune $\lambda$ during both token-level sampling within sequences and across multiple sampled sequences. See \autoref{app:adaptive_lambda} for details of the adaptive tuning procedure.

\subsection{Fine-tuning NLI for partial sequences} \label{sec:finetune_nli}

Standard NLI models expect full premise–hypothesis pairs, but our sampler requires entailment scores for partially generated or masked text. To adapt an off‐the‐shelf NLI model with minimal overhead, we begin by loading a model finetuned on a natural language understanding dataset such as the MNLI benchmark dataset \citep{williams2018broad} (\textit{e.g.}, DeBERTa‐large‐MNLI \citep{he2020deberta}) and freezing all of its existing parameters. We then consider a special token: \texttt{[TRUNC]}\footnote{If the token does not exist in the tokenizer's vocabulary, we add it and initialize its embedding randomly.} in the case of ARMs and \texttt{[MASK]} in the case of MDMs. The token embedding and the model’s final classification layer are the only components allowed to update during finetuning.

\begin{figure}[t]
    \centering
    \includegraphics[width=\linewidth]{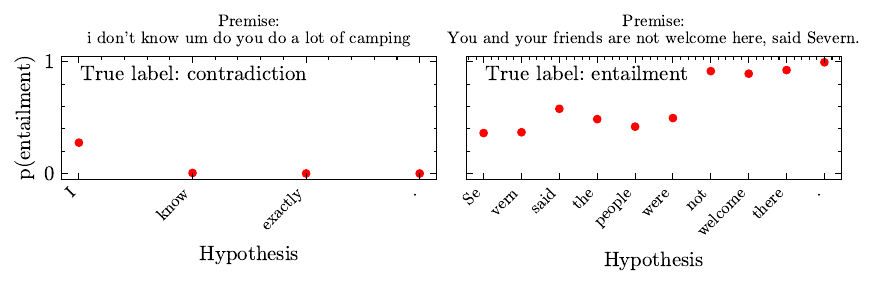}
    \caption{Predictions of the finetuned NLI at each truncated point of the hypothesis, on two examples from the GLUE MNLI \texttt{validation\_matched} split \citep{wang2018glue}.}
    \label{fig:nli_trunc} 
\end{figure}

Next, we construct an augmented dataset by corrupting exactly one side of each NLI training example. The corruption mechanism is \textit{truncation} for ARMs and \textit{masking} for MDMs. For each premise–hypothesis pair, we randomly determine whether to corrupt the premise or the hypothesis. For ARMs, we truncate the selected sequence of length $T$ at a point drawn uniformly at random from $t \in \{1, \dotsc, T\}$ and append the token \texttt{[TRUNC]} if $t<T$. This procedure exposes the model to cases where either the premise or the hypothesis ends abruptly, teaching it to interpret \texttt{[TRUNC]} as a signal of incompleteness. For MDMs, we randomly select the masking probability uniformly in $[0, 1]$ and independently replace tokens with \texttt{[MASK]} with that probability.

We then finetune using the standard cross entropy loss, updating only the \texttt{[TRUNC]} or \texttt{[MASK]} embedding, \texttt{[CLS]}\footnote{Keeping the \texttt{[CLS]} frozen had little effect on performance.} embedding, and the classification head, which corresponds to only 0.3\% of the model parameters (around 3M parameters) in the case of DeBERTa‐large‐MNLI \citep{he2020deberta}. This adaptation is thus lightweight and preserves the model’s original NLI performance. At inference time, for ARMs, whenever we need to score a partial prefix against a full previous sample (or vice versa), we append \texttt{[TRUNC]} to the truncated side and query the finetuned NLI model for the probability of entailment. For MDMs, no preprocessing is needed to query the finetuned NLI model, as its vocabulary already contains the \texttt{[MASK]} token. 

\begin{figure}
    \centering
    \includegraphics[width=\linewidth]{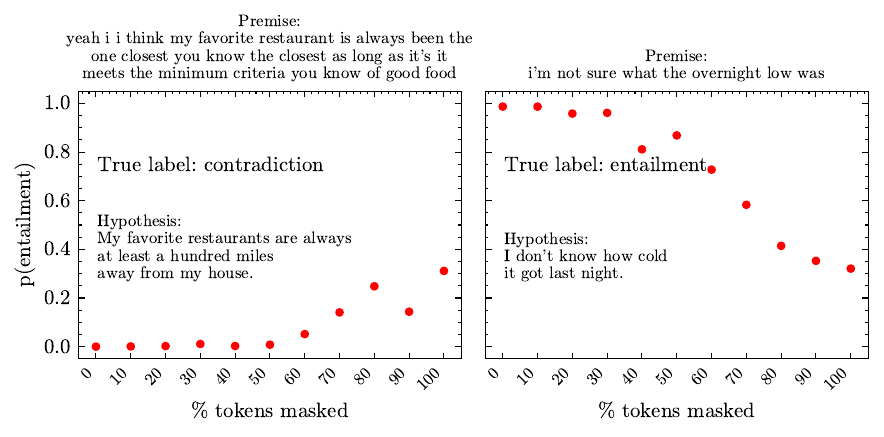}
    \caption{Predictions of the finetuned NLI at each masking percentage of the hypothesis, on two examples from the GLUE MNLI \texttt{validation\_matched} split \citep{wang2018glue}.}
    \label{fig:nli_mask}
\end{figure}

\autoref{fig:nli_trunc} traces the probability that the fine-tuned NLI model assigns to the ``entailment'' class as progressively longer prefixes of each hypothesis are revealed. In both panels the trajectory converges to the ground-truth label long before the final token appears, suggesting that even partially generated hypotheses already contains enough semantic signal for identifying entailment. See \autoref{fig:nli_mask} for a similar demonstration for the MDM case. In another view, \autoref{fig:accuracy_vs_corruption} plots the classification accuracy of the finetuned NLI model at varying corruption levels. For both ARMs and MDMs, the classification accuracy matches that of the NLI model prior to finetuning and slowly falls until it reaches the ``random guess'' accuracy at complete corruption.


\subsection{Importance-reweighted estimators of uncertainty} \label{sec:importance_weighted_estimators}
As we sample from the biased proposal $q(\cdot)$ in \cref{eq:modified_dist_ar,eq:modified_dist_mdm} instead of the true model distribution $p(\cdot)$, we apply importance reweighting to correct for the introduced bias. For a set of $N$ generated sequences $\{s^{(i)}\}_{i=1}^N$ where each $s^{(i)}$ was drawn from $q(s^{(i)})$, we compute the unnormalized and self-normalized importance weights:
\begin{equation} \label{eq:importance_weights}
  w_i \;=\;\frac{p\bigl(s^{(i)}\bigr)}{q\bigl(s^{(i)}\bigr)}, 
  \quad
  \tilde w_i \;=\;\frac{w_i}{\sum_{j=1}^N w_j}\,.
\end{equation}
In this section, we illustrate how importance reweighting and semantic clustering interact for the purpose of estimating SE \citep{kuhn2023semantic} and MI \citep{yadkori2024believe}, proxies of aleatoric and epistemic uncertainty, respectively. Note that these are meant to serve as illustrating examples and we do not advocate for these particular uncertainty metrics over alternatives.

\paragraph{Semantic (aleatoric) uncertainty.}
Let $\mathcal C$ be the set of semantic clusters. The probability of each cluster $c \in \mathcal{C}$ can be estimated using the importance-weighted version of \autoref{eq:cluster_prob_mc}:
\begin{equation}
  \hat p(c) \approx \sum_{i=1}^N 1[s^{(i)} \in c] \ \tilde w_i. 
\end{equation}
This yields the importance-weighted version \citep{aichberger2025improving} of SE \citep{kuhn2023semantic} in \autoref{eq:semantic_entropy}:
\begin{equation}
  \label{eq:imp_semantic_entropy}
  \hat H =
  -\sum_{c\in\mathcal C}\hat p(c)\,\log\hat p(c).
\end{equation}

\paragraph{Epistemic uncertainty.}
We use the MI proxy introduced by \citet[Algorithm 2]{yadkori2024believe}, restricted here to the pairwise case $n{=}2$, as a worked example of how one can quantify epistemic uncertainty with our semantically–diverse sample set. Rather than a faithful reproduction of the original MI implementation, we aim at demonstrating our diversity sampling within an importance sampling framework. The MI proxy operates on the assumption that successive answers are \textit{conditionally independent} under the ground-truth distribution $p^*$:
$$
p^*(y_1,y_2\mid x) \;=\; p^*(y_1\mid x)\,p^{\!*}(y_2\mid x)
$$
so that any observed dependence between them can be ascribed to epistemic uncertainty.

We draw $N$ pairs $z^{(i)} \coloneqq (s^{(i)}_1,s^{(i)}_2)$ from the diversity-steered distribution $q^{e}$ by iterative prompting. Each response token sequence is mapped to a semantic cluster $c^{(i)}_j{\in}\mathcal C$ using the NLI-reported bidirectional entailment score \citep{kuhn2023semantic}, rather than using the F1 score as originally done in \citet{yadkori2024believe}. 
We find cluster centers $\mathcal{S}$ defined by all tuples $j, k$ such that $E(z^{(j)}, z^{(k)}) < \tau$ and each cluster center $z^\star \in \mathcal{S}$ is associated with a cluster $c(z^\star) \coloneqq \{z': E(z^\star, z') \geq \tau \}$.

While we sample from the proposal $q^{e}$, the target distribution for MI is the model distribution $p^{e}$, so each pair receives a self-normalized weight
$$
w_{i, \textrm{joint}} \;=\; \frac{p^{e}\!\bigl(s^{(i)}_1,s^{(i)}_2 \bigr)}{q^{e}\!\bigl(s^{(i)}_1,s^{(i)}_2 \bigr)},
\qquad
\tilde w_{i, \textrm{joint}} \;=\; \frac{w_{i, \textrm{joint}}}{\sum_{j=1}^{N} w_{j, \textrm{joint} }}\, .
$$

After clustering the response pairs, we evaluate the joint and marginal cluster probabilities:
\begin{align} \label{eq:cluster_probs_mi}
\hat p(c(z_1, z_2)) &=
\sum_{i=1}^{N}
1[(s^{(i)}_1, s^{(i)}_2) \in c(z_1, z_2) \bigr] {\tilde w}_{i, \textrm{joint}},
\end{align}

\begin{align}
\hat p^{\otimes}\!\bigl(c(z_1,z_2)\bigr)
   \;=\;
   \Bigl[\!
        \sum_{z' \in \mathcal{S}}
        \hat p\bigl(c(z_1, z'_2)\bigr)
   \Bigr]
   \;
   \Bigl[\!
        \sum_{z'' \in \mathcal{S}}
        \hat p\bigl(c(z''_1, z_2)\bigr)
   \Bigr].
\end{align}

Finally, the importance-reweighted MI estimator becomes:
\begin{equation} \label{eq:imp_mi_estimator}
\hat I =
\sum_{c \in \mathcal C}
\hat p(c)\,
\log\!\frac{\hat p(c)}
           {\hat p^\otimes(c)} .
\end{equation}
A large $\hat I$ signals strong dependence between successive clustered answers (hence high epistemic uncertainty) whereas $\hat I\!\approx\!0$ is consistent with the conditional independence assumption and suggests that the model’s current parameters are locally trustworthy for the given query. 

\subsection{Variance reduction via adaptive control variates}
The importance-weighted estimators for SE and MI can exhibit high variance, particularly when the proposal $q(\cdot)$ differs substantially from the model distribution $p(\cdot)$.  To mitigate this, we employ \emph{control variates}; we choose a proxy statistic correlated with our target and estimate its coefficient from the same weighted samples. They may optionally be applied \textit{adaptively} on the running samples. \autoref{app:variance_reduction} provides additional background and derivations.

\paragraph{Semantic entropy.}
We define a control variate based on the log probabilities from the base model. Denote $Y_i = -\log \hat{p}(c(s^{(i)}))$, where $c(s^{(i)})$ is the cluster containing $s^{(i)}$. The estimator in \autoref{eq:imp_semantic_entropy} is then $\sum_{i=1}^N \tilde w_i Y_i$. Now define $X_i = -\log p(s^{(i)})$, and let $X_i' = X_i - \mu_X$ and $Y_i' = Y_i - \mu_Y$ represent the centered versions, with empirical means $\mu_X = \sum_{i=1}^N \tilde w_i X_i$ and $\mu_Y = \sum_{i=1}^N \tilde w_i Y_i$. Our adjusted entropy estimator is then:
\begin{equation} \label{eq:H_cv}
  \hat{H}_{\textrm{cv}} = \underbrace{\sum_{i=1}^N \tilde w_i Y_i}_{\hat H \textrm{ in Eq. (10)}} - \alpha \underbrace{ \sum_{i=1}^N \tilde w_i X_i'}_{\textrm{control variate}},
\end{equation}
where the optimal coefficient $\alpha$ that minimizes variance is computed as:
\begin{equation*}
\alpha^\star_{\textrm{SE}} = \frac{\sum_{i=1}^N \tilde w_i X_i' Y_i'}{\sum_{i=1}^N \tilde w_i {X_i'}^2}.
\end{equation*}
This scheme leverages the correlation between the log model probabilities and the cluster entropy to reduce variance without extra inference cost; note that $X_i$ is already required to compute $w_i$.

\paragraph{Mutual information.}
Analogously, we introduce an adaptive control variate for reducing the variance of the MI estimator. Letting $Y_i = \log (\hat p(c(z_1,z_2)) / \hat p^\otimes(c(z_1,z_2)))$ represent the log ratio appearing in \autoref{eq:imp_mi_estimator}, we similarly define the control variate using the joint log probability under the base model, $X_i = \log p(s^{(i)}_1, s^{(i)}_2 )$, with means and centered variables defined analogously to the SE case. Our control variate-corrected estimator is thus:
\begin{equation} \label{eq:I_cv}
  \hat{I}_{\text{cv}} = \sum_{i=1}^N \tilde w_{i,\textrm{joint}} Y_i - \alpha_{\textrm{MI}}\sum_{i=1}^N \tilde w_{i,\textrm{joint}} X_i',
\end{equation}
with the adaptively computed coefficient given by:
\begin{equation*}
\alpha^\star_{\textrm{MI}} = \frac{\sum_{i=1}^N \tilde w_{i,\textrm{joint}} X_i' Y_i'}{\sum_{i=1}^N \tilde w_{i,\textrm{joint}} {X_i'}^2}.
\end{equation*}

\section{Related Work}

\paragraph{Uncertainty estimation.} Common approaches for uncertainty quantification in vision and classification tasks include Monte Carlo dropout \citep{gal2016dropout}, deep ensembles \citep{lakshminarayanan2017simple}, and prior networks \citep{malinin2018predictive, malinin2020ensemble}. While these methods have also been adapted for text classification and regression in NLP \citep{jiang2018trust, desai2020calibration, wang2022uncertainty,wiegreffe-etal-2023-increasing}, their extension to free-form NLG presents specific challenges due to semantic invariances inherent in generated text \citep{glushkova2021uncertainty,kuhn2023semantic}.
Several studies have proposed prompting or fine-tuning language models to explicitly articulate their confidence levels \citep{kadavath2022language,mielke2022reducing,cohen2023crawling,Ganguli:23,pmlr-v239-ren23a,Tian:23,hager2025uncertainty}, though this typically requires additional supervision. Attention values may contain information about relevance or confidence \citep{lin-etal-2024-contextualized,duan-etal-2024-shifting}. Alternatively, cross-examination leverages a secondary language model to evaluate uncertainty in another model's outputs \citep{cohen2023crossexam}. Predictive entropy quantifies the token-level entropy of the predictive distribution \citep{malinin2018predictive}. Semantic entropy \citep{kuhn2023semantic} offers an unsupervised method that clusters multiple outputs into semantic equivalence classes based on bidirectional entailment and then computes entropy on these clusters. Complementary strategies using the conformal framework can provide bounds on errors under stronger theoretical assumptions \citep{ravfogel2023conformal, angelopoulos2024conformal}.

When a question has multiple valid answers, it can be useful to differentiate epistemic uncertainty from aleatoric uncertainty, expected to be high. Classically, the former is defined by an expectation over possible weight realizations (\autoref{eq:semantic_epistemic_bayesian}), which vanishes when all admissible weight vectors agree. Recent work, however, suggests that one can probe it with a \textit{single fixed} network by measuring the \emph{consistency} of multiple answers it produces to the same query \citep{ahdritz2024distinguishing,johnson2024experts,yadkori2024believe}. 

\paragraph{Diversity-promoting sampling.} Sampling heuristics aimed at enhancing output diversity, such as temperature scaling, top-$k$, and nucleus sampling \citep{holtzman2019curious} do not account for semantics. 
Diverse beam search (DBS) introduces diversity heuristics within beam search optimization \citep{Vijayakumar:18}. Contrastive decoding enlists a secondary, weaker language model whose output tokens are penalized to encourage diverse token selection by the primary model \citep{Li2023contrastive}. Cluster-based beam search methods apply semantic clustering to prune beam candidates and diversify subsequent selections \citep{li2016diversity,tam2020clusterbeam}, but this heavily depends on initial candidate diversity. Semantically Diverse Language Generation (SDLG) substitutes the most informative token in a fully generated sample and allows standard sampling to proceed from that token onward \citep{aichberger2025improving}. It does not, however, explicitly account for diversity within a running sample set and, because the token scoring involves a gradient of the NLI loss with respect to the NLI token space, requires additional implementation \citep[\textit{e.g.},][]{artetxe-etal-2016-learning} if the base LLM does not share the tokenization scheme or vocabulary with the NLI. Our approach differs by integrating a continuous semantic penalty directly into the sampling logits during text generation in a gradient-free manner. For comparative analyses of diversity-enhancing decoding strategies, see \citet{ippolito2019comparison}.

\paragraph{Semantic clustering and paraphrase detection.} Grouping model outputs based on semantic equivalence often reduces to paraphrase detection. Casting semantic equivalence as bidirectional entailment dates back to early linguistics work \citep{culicover1968paraphrase} and was later adopted in NLP \citep{pado2009machine,androutsopoulos2010survey,pavlick2015ppdb}. Early methods relied on lexical overlap \citep{qiu2006paraphrase} or vector embedding similarities \citep{socher2011dynamic,yu2015learning}. \cite{bannard2005paraphrasing} explored the use of bilingual parallel corpora for paraphrase extraction and ranking. BERT-style encoders can be used to build binary ``paraphrases'' vs. ``not paraphrases'' classifiers \citep{devlin2019bert,phrasebertwang2021}. The entailment probability from NLI models has been used to cluster LLM generations into meaning-equivalent sets, enabling unsupervised uncertainty estimation \citep{kuhn2023semantic,aichberger2025improving,lin2023generating}.

\section{Experiments} \label{sec:experiments}

We evaluate our diversity‑promoting sampling scheme for estimating SE \citep{kuhn2023semantic} and MI \citep{yadkori2024believe}. The target estimators are the importance-reweighted versions with control variates: $\hat H_{\rm cv}$ in \autoref{eq:H_cv} and $\hat I_{\rm cv}$ in \autoref{eq:I_cv}, respectively. Due to space constraints, implementation details, additional results for $\hat H$ including ablation studies, and results for $\hat I_{\rm cv}$ are deferred to \Cref{app:implementation,app:semantic_entropy,app:mutual_info}. As observed by \citet{aichberger2025improving}, it is common to employ diversity heuristics during sampling without correcting for the introduced bias with principled importance reweighting \citep[\textit{e.g.},][]{kuhn2023semantic,yadkori2024believe}. 
Although one of our contributions is the variance-reduced application of importance correction with \textit{control variates}, we apply the \textit{same} estimation procedure to the final generations from all sampling schemes, to enable a fair comparison. Estimation is performed on top of the semantic clusters created using \citet[Algorithm 1]{kuhn2023semantic}. In brief, clustering works by querying the DeBERTa-large-MNLI model \citep{he2020deberta}, finetuned on the NLI dataset MNLI \citep{williams2018broad}, on every pair of sampled responses. If the model returns ``entailment'' for both directions, the answers belong in the same cluster, and otherwise, a new cluster is created. 

\paragraph{Sampling baselines.}
 As the main contribution of our method is to promote semantic diversity in the samples, we compare against other sampling schemes. Our ARM baselines include (1) standard IID sampling with temperatures $\tau \in \{1, 2\}$, (2) diverse beam search (DBS) \citep{Vijayakumar:18} with a penalty hyperparameter of 0.5, and (3) our re-implementation of SDLG  \citep{aichberger2025improving}, where we handle the differing vocabularies of the base LLM and NLI by decoding each OPT/LLaMA token to its raw string and then re‐tokenizing that string with the DeBERTa NLI tokenizer, so that substitution candidates and gradient attributions live in the same NLI embedding space. For the MDM, we compare with Gumbel temperatures of $\tau \in \{1, 2\}$.

\begin{table}[ht]
\centering
\small
\caption{AUROC of SE \citep{kuhn2023semantic} computed on generations from various sampling schemes. Each scheme uses $N{=}16$ sequences. The correctness metric Rouge-L (F1 score) was thresholded at 0.3. All numbers are ${\rm mean}{\pm}{\rm std}$ over 5 jackknife samples of size 200. The symbol ``$-$'' indicates that the sampling scheme does not apply to MDMs. ``Vanilla'' refers to standard sampling without any tempering. Best methods based on mean are bolded. For AmbigQA, we omit results for OPT-6.7B, as it generated a high fraction of nonsensical responses. \label{tab:se_results}}
\begin{tabular}{llccccc}
\toprule
\textbf{Dataset} & \textbf{Model} & {Vanilla ($\tau=1$)} & {$\tau=2$} & {DBS \citep{Vijayakumar:18}} & {SDLG \citep{aichberger2025improving}} & {Ours} \\
\midrule \midrule
\multirow{4}{*}{CoQA} 
  & OPT-6.7B  & .59${\pm}$.06 & .69${\pm}$.04 & .68${\pm}$.04 & .71${\pm}$.02 & \textbf{.75${\pm}$.02} \\
  & OPT-13B  & .70${\pm}$.04 & \textbf{.76${\pm}$.04} & .73${\pm}$.04 & .72${\pm}$.02 & .75${\pm}$.03 \\
  & LLaMA 3 8B-Instruct   & .68${\pm}$.03 & .72${\pm}$.04 & .71${\pm}$.05 & .74${\pm}$.02 & \textbf{.77${\pm}$.02} \\
  & LLaDA 8B-Instruct   & .78${\pm}$.02 & \textbf{.81${\pm}$.05} & - & - & \textbf{.81${\pm}$.04} \\
\midrule
\multirow{4}{*}{TriviaQA} 
  & OPT-6.7B  & .66${\pm}$.05 & .67${\pm}$.06 & .71${\pm}$.04 & .78${\pm}$.03 & \textbf{.82${\pm}$.03} \\
  & OPT-13B  & .72${\pm}$.04 & .70${\pm}$.05 & .73${\pm}$.04 & \textbf{.86${\pm}$.03} & .85${\pm}$.03 \\
  & LLaMA 3 8B-Instruct   & .79${\pm}$.04 & .70${\pm}$.04 & .70${\pm}$.03 & .79${\pm}$.04 & \textbf{.84${\pm}$.03} \\
  & LLaDA 8B-Instruct   & .81${\pm}$.11 & .83${\pm}$.05 & - & - & \textbf{.86${\pm}$.04} \\
\midrule
\multirow{4}{*}{AmbigQA} 
  & OPT-13B  & .65${\pm}$.10 & .68${\pm}$.11 & \textbf{.78${\pm}$.08} & .71${\pm}$.08 & \textbf{.78${\pm}$.04} \\
  & LLaMA 3 8B-Instruct   & .70${\pm}$.04 & .55${\pm}$.07 & .71${\pm}$.08 & \textbf{.77${\pm}$.05} & .76${\pm}$.03 \\
  & LLaDA 8B-Instruct   & .70${\pm}$.09 & .71${\pm}$.08 & - & - & \textbf{.76${\pm}$.03} \\
\midrule
\multirow{4}{*}{TruthfulQA} 
  & OPT-6.7B  & .80${\pm}$.04 & .80${\pm}$.05 & .77${\pm}$.02 & .78${\pm}$.06 & \textbf{.81${\pm}$.06} \\
  & OPT-13B  & .73${\pm}$.06 & .74${\pm}$.08 & .79${\pm}$.05 & .81${\pm}$.04 & \textbf{.85${\pm}$.04} \\
  & LLaMA 3 8B-Instruct   & .88${\pm}$.04 & .88${\pm}$.05 & \textbf{.89${\pm}$.04} & .86${\pm}$.04 & \textbf{.89${\pm}$.02} \\
  & LLaDA 8B-Instruct   & .85${\pm}$.04 & .89${\pm}$.04 & - & - & \textbf{.94${\pm}$.02} \\
\bottomrule
\end{tabular}
\end{table}

\paragraph{Datasets.}
We perform experiments on four question-answering (QA) benchmark datasets covering both closed-book and open-book tasks: 907 validation matched instances with shorter stories from \textbf{CoQA} \citep{reddy2019coqa}, a closed-book abstractive QA; 1,000 instances from the validation no-context reading comprehension split of \textbf{TriviaQA} \citep{joshi2017triviaqa}, a closed-book extractive QA; 800 instances from the validation split of \textbf{TruthfulQA} \citep{lin-etal-2022-truthfulqa}, a closed-book generative QA; and the light validation split of \textbf{AmbigQA} \citep{min2020ambigqa}, an open-book open-domain QA. Because AmbigQA contains multi-answer questions requiring question disambiguation and abstractive responses, it serves as a test environment with highly ambiguous questions, where $\hat H_{\rm cv}$ is expected to be large. 

\paragraph{Models.} We apply our method to four models spanning a range of QA capabilities as well as both ARM and MDM sampling paradigms: \textbf{OPT‑6.7B}, \textbf{OPT‑13B} \citep{zhang2022opt}\footnote{We omit OPT-30B, as the marginal AUROC improvement relative to OPT-13B has been insignificant \citep{aichberger2025improving,duan-etal-2024-shifting}.} for comparisons with prior work \citep{kuhn2023semantic,aichberger2025improving,duan-etal-2024-shifting}; \textbf{LLaMA 3 8B-Instruct} \citep{touvron2023llama,llama3modelcard}, as a modern instruction-finetuned backbone; and \textbf{LLaDA 8B-Instruct} \citep{nie2024llada}, an instruction-finetuned MDM.

\paragraph{Metrics.} Following the evaluation procedures in prior work, we report the Area Under the Receiver Operating Characteristic curve (AUROC), where the correct answer is defined by ROUGE‑L (F1 score) $<0.3$ against the reference answer. When there are more than one reference answers, we take the maximum score across the answers. We also evaluate the average number of clusters and the effective sample size ($\rm ESS$) \citep{kong1992note} of the importance weights relative to $N{=}16$.

\section{Discussion and limitations} \label{sec:discussion}
\begin{wrapfigure}{r}{0.35\textwidth}
    \centering
    \vspace{-0.4cm}
    \includegraphics[width=0.34\textwidth]{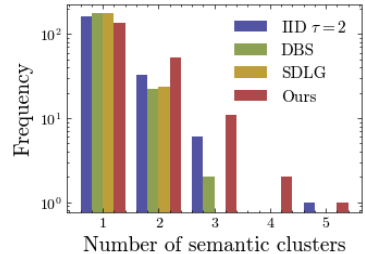}
    \caption{Number of semantic clusters captured by various sampling schemes on the CoQA dataset.\vspace{-0.2cm}}
    \label{fig:num_clusters}
\end{wrapfigure}
By design, our sampling scheme covers more semantic clusters than the baselines while using the same number of samples (see \autoref{fig:num_clusters}).
Meanwhile, applying importance correction with control variates preserves the rank agreement between estimated uncertainties and correctness, as reflected in competitive or superior AUROC across models and datasets (see \autoref{tab:se_results}). Consistent with \citet{aichberger2025improving}, we find that semantics-agnostic heuristics such as simple temperature scaling or DBS \citep{Vijayakumar:18} are insufficient to fully explore the semantic space. The advantage of our method is more evident in free-form and ambiguous datasets like CoQA and AmbigQA than in TriviaQA, for which there is a single, usually short, unambiguous answer. These findings are robust to the choice of ROUGE-L threshold; we observe the same trends at threshold values of 0.1 and 0.5 (\autoref{app:semantic_entropy}). As a complementary threshold-free metric, the Spearman $\rho$ between the negated ROUGE-L scores and our estimated uncertainties were 5\% and 6\% greater on average than those of DBS and SDLG, respectively. We suspect SDLG’s weaker rank correlation may be due to our simplified handling of differing tokenizations between OPT/LLaMA and the NLI model. Finally, the ratio ${\rm ESS}/N$ stays above 0.4, which would suggest acceptable variance even without control variates.

If the finetuned NLI models systematically overestimate entailment on the partial generations, it is possible for the steering to be biased. We empirically observe, however, that the predicted entailment probability sharply peaks at the ``random guess'' value of $1/3$ when the input generation has one token revealed (\autoref{fig:entail_prob_minimal_info}). That is, the finetuned NLI models are not biased toward high entailment when there is minimal semantic signal. The value of $1/3$ motivates the choice of our default schedule for $\lambda$ (\autoref{app:adaptive_lambda}), where we increase $\lambda$ when the bidirectional entailment score of the current generation with respect to the most similar existing generation is higher than $1/3$.

In general, sampling algorithms face a trade-off between encouraging \textit{exploration} by introducing a joint steering term on the walkers and facilitating \textit{parallel sampling} by preserving each walker’s independence. Our method can be viewed as prioritizing the former; we enforce diversity sequentially in the running sample set \citep{search_and_eval}, at the expense of computation time that grows linearly with $N$. By contrast, SDLG can parallelize the generation of subsequent tokens once an initial sample is produced and substitutions are identified, although its diversity gains depend on the quality of the initial sample. Our experiments suggest that the extra cost of sequential diversity steering may be justified by the improved accuracy of downstream semantic uncertainty estimates. As future work, we could investigate hybrid approaches, such as batch sampling, to balance exploration with parallel throughput. Note also that, although we fixed $N$ for all sampling schemes here, the online stopping mechanism based on estimator stability, described in \autoref{app:implementation}, can help with sample efficiency. 

\section{Conclusion} \label{sec:conclusion}
We presented \emph{diversity-steered sampling}, a simple plug-in that adds a bidirectional-entailment repulsion term to both ARM and MDM decoding, and then corrects the resulting bias with importance weights to recover consistent estimates of SE (aleatoric uncertainty) and a MI lower bound on epistemic uncertainty. Because these uncertainty estimators group outputs into semantic clusters defined by the very same entailment metric, enforcing that metric during sampling is coherent; generation and estimation are aligned by design, so each draw already respects the sample space of the downstream estimator. To our knowledge, this is the first framework that (1) applies to both ARM and MDM decoding paradigms (including recent models such as LLaMA-v3 \citep{llama3modelcard} and LLaDA \citep{nie2024llada}), (2) has been demonstrated on both aleatoric and epistemic proxies, and (3) requires no gradient access to the NLI or the base LLM.

Several opportunities for improving robustness remain. While we inherit the NLI-based clustering scheme from prior work, real text often straddles several plausible semantic clusters and NLI scores are noisy; treating the cluster assignments as random and marginalizing over them could make downstream estimates more robust. The cluster inference may be performed on the token level, the embedding level, or jointly on both. Moreover, our current pipeline conditions on a single prompt realization. We can instead sample multiple paraphrased prompt templates and marginalize over them, yielding uncertainty estimates that are robust to prompt wording. Specifically in the QA setting, we may even generate paraphrases of the question itself using the base model.

Looking ahead, the same logit–repulsion plus self-normalized importance-weighting scheme can, in principle, be embedded inside on-policy RL fine-tuning methods such as Proximal Policy Optimization (PPO) \citep{DBLP:journals/corr/SchulmanWDRK17} and its grouped-reward variant GRPO \citep{guo2025deepseek}.
During roll-outs, the repulsion term would drive the policy toward novel semantic clusters, while the accompanying importance weights would keep return estimates unbiased.



\section*{Acknowledgments}
We thank Aya Ismail for helpful discussions on MDM decoding. 

\medskip
\newpage
\bibliography{main}
\bibliographystyle{neurips}

\newpage
\appendix
\section*{Technical Appendices and Supplementary Material}

\section{Extension to masked diffusion models} \label{app:mdm}

Algorithm~\ref{alg:diversity_sampling_mdm} provides a simplified pseudocode of our diversity steering scheme for MDMs, assuming a denoising schedule where one masked token is predicted at a time. \autoref{fig:nli_mask} plots the entailment probability against the fraction of hypothesis tokens that are randomly masked. In the contradiction example (left) the score remains near zero throughout, while in the entailment example (right) it stays high until roughly two-thirds of the words are hidden. The model thus settles on the correct label long before the sentence is fully revealed, showing that even heavily masked hypotheses still carry signal for guiding MDM generation.

\begin{algorithm}
\small
\caption{Diversity‐steered masked‐diffusion sampling}\label{alg:diversity_sampling_mdm}
\begin{algorithmic}[1]
\Require Prompt $x$; base masked‐diffusion model $p(\cdot \mid \cdot)$; bidirectional NLI scorer $E(\cdot,\!\cdot)$ from \autoref{eq:bidirectional_entailment}, trained with a special marker $\texttt{[MASK]}$ for incomplete spans; diversity penalty $\lambda$; number of samples $N$; total denoising steps $T$
\Ensure Set of semantically diverse generations $\mathcal{S}$
\State $\mathcal{S} \gets \varnothing$
\For{$i = 1$ \textbf{to} $N$}
  \State $y^{(T)} \gets \textsc{MaskTokens}(x)$ 
  \Comment{Initialize by masking random spans of the input sequence}
  \For{$t = T{-}1$ \textbf{downto} $0$}
      \State $\mathcal{M}_t \gets \textsc{SelectMask}(y^{(t+1)})$ 
      \Comment{Choose subset of masked positions to fill at step $t$}
      \For{each position $m \in \mathcal{M}_t$}
          \For{each candidate token $\mathrm{next} \in \mathcal{V}$} \Comment{Alternatively consider the top-$k$ tokens only}
              \State $\ell(\mathrm{next}) \gets \log p\bigl(y^{(t)}_m{=}\mathrm{next} \mid y^{(t+1)}\bigr)$ 
              \Comment{Base model logits for current mask}
              \State $\hat y^{(t)} \gets y^{(t+1)}$;  $\hat y^{(t)}_m \gets \mathrm{next}$
              \Comment{Temporarily fill mask with candidate token}
              \State $\hat s \gets \mathrm{decode}(\hat y^{(t)})$
              \Comment{Form partially denoised sequence}
              \State $\pi \gets \displaystyle\max_{s \in \mathcal{S}} E\bigl(\hat s, s\bigr)$
              \Comment{Similarity score with the most similar existing generation}
              \State $\ell'(\mathrm{next}) \gets \ell(\mathrm{next}) - \lambda \pi$
              \Comment{Repel toward semantic novelty}
          \EndFor
          \State $y^{(t)}_m \sim \mathrm{Categorical}\bigl(\mathrm{softmax}(\ell')\bigr)$
          \Comment{Sample a token for the current masked position}
      \EndFor
      \State $y^{(t)} \gets \textsc{FillMasks}(y^{(t+1)}, y^{(t)}_{\mathcal{M}_t})$
      \Comment{Update denoised sequence}
  \EndFor
  \State $s^{(i)} \gets \mathrm{decode}(y^{(0)})$
  \State $\mathcal{S}\!\leftarrow\!\mathcal{S}\,\cup\,\{s^{(i)}\}$
\EndFor
\State \Return $\mathcal{S}$
\end{algorithmic}
\end{algorithm}

\begin{figure}[htbp]
    \centering
    \begin{subfigure}[b]{0.44\textwidth}
        \includegraphics[width=\textwidth]{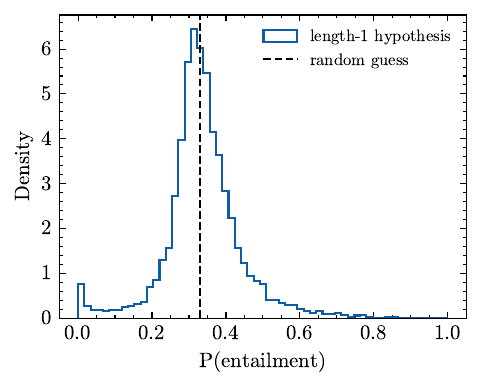}
        \caption{Prefix truncation}
        \label{fig:entail_prob_minimal_info_arm}
    \end{subfigure}
    \hfill
    \begin{subfigure}[b]{0.46\textwidth}
        \includegraphics[width=\textwidth]{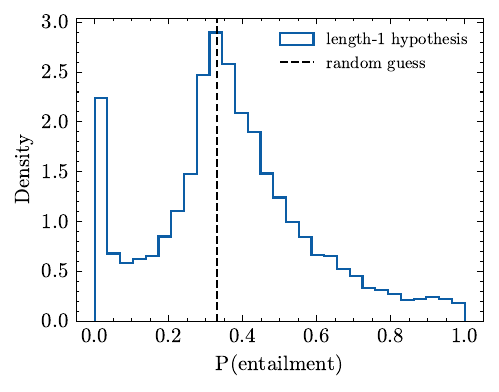}
        \caption{Masked tokens}
        \label{fig:entail_prob_minimal_info_mdm}
    \end{subfigure}
    \caption{Distributions of entailment probability predicted by the finetuned NLI on the GLUE MNLI \texttt{validation\_matched} dataset when (a) truncating or (b) masking all but one token of the hypothesis. Both distributions are peaked at the ``random guess'' probability of $1/3$.}
    \label{fig:entail_prob_minimal_info}
\end{figure}

\begin{figure}[htbp]
    \centering
    \begin{subfigure}[b]{0.45\textwidth}
        \includegraphics[width=\textwidth]{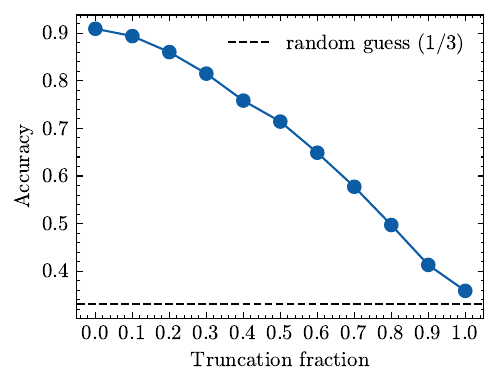}
        \caption{Prefix truncation}
        \label{fig:accuracy_vs_truncation}
    \end{subfigure}
    \hfill
    \begin{subfigure}[b]{0.45\textwidth}
        \includegraphics[width=\textwidth]{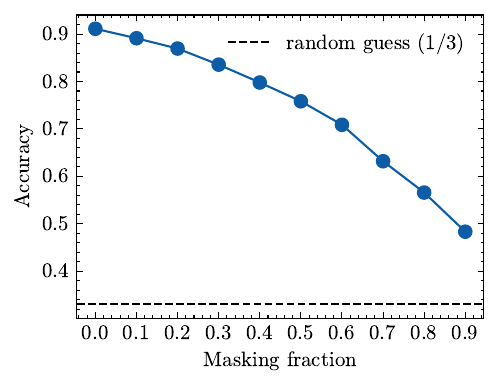}
        \caption{Masked tokens (averaged over 20 realizations)}
        \label{fig:accuracy_vs_masking}
    \end{subfigure}
    \caption{Classification accuracy of finetuned NLI models at varying corruption levels for (a) prefix truncation and (b) masked tokens. At zero corruption, the accuracy matches that of the pretrained model and slowly falls until it reaches the ``random guess'' accuracy of $1/3$ at complete corruption.}
    \label{fig:accuracy_vs_corruption}
\end{figure}

\section{Additional implementation details} \label{app:implementation}

Our sampling pipeline is implemented at \href{https://github.com/jiwoncpark/diversity_steered_sampling}{\url{https://github.com/jiwoncpark/diversity\_steered\_sampling} \faGithub}.

\subsection{NLI finetuning} \label{app:nli}

\paragraph{Prefix truncation.}
We start from the \texttt{microsoft/deberta-large-mnli} checkpoint corresponding to the DeBERTa model \citep{he2020deberta} finetuned on the GLUE Multi-NLI matched dataset \citep{wang2018glue}. To mark a sequence as partially generated, we introduce a dedicated \texttt{[TRUNC]} token. The MNLI dataset is augmented as follows. Every MNLI pair is deterministically unrolled into (i) the original sentence pair, (ii) all proper prefixes of the hypothesis (lengths $1\ldots L_{\!h}-1$) each followed by \texttt{[TRUNC]}, and (iii) all proper prefixes of the premise (lengths $1\ldots L_{\!p}-1$) similarly tagged. A single example with tokenized lengths $(L_{\!p},L_{\!h})$ thus contributes $1+(L_{\!h}-1)+(L_{\!p}-1)$ training instances, covering every possible truncation. All backbone model parameters are frozen, but three small modules remain trainable: the single embedding row for \texttt{[TRUNC]}, the full classification head, and the pooler projection. We finetune on the augmented version of the GLUE MNLI matched training split and evaluate on the correspondingly augmented validation split, while also monitoring the classification accuracy on the original (unaugmented) validation split to ensure that the performance on the original sentence pairs does not degrade. Optimization uses \textsc{AdamW} \citep{loshchilov2017decoupled} (initial learning rate~$5{\times}10^{-5}$, weight decay~$0.01$) with a batch size of 8 for two epochs. The final validation accuracy on the augmented set was 73.3\% and that on the original set was 91.0\%, which was similar to the accuracy prior to finetuning (90.8\%).

\paragraph{Masked tokens.}
Starting from the same \texttt{microsoft/deberta-large-mnli} checkpoint, we first ensure that the tokenizer exposes a \texttt{[MASK]} token. During training, each MNLI example is expanded on-the-fly into one intact pair plus $20$ stochastic variants in which either the premise or the hypothesis has a uniformly random fraction of tokens ($f\!\sim\!\mathcal{U}(0,1)$) replaced by \texttt{[MASK]}. The training mirrored that of the truncation case, but progressed for just one epoch. The final validation accuracy on the augmented set was 73.9\% and that on the original set was 91.1\%, which was similar to the accuracy prior to finetuning (91.2\%).

\subsection{Prompts} \label{app:prompts}
We use the following prompt for the SE experiments on CoQA \citep{reddy2019coqa}.
\begin{verbatim}
<context> Answer in one sentence. Q: <question> A: 
\end{verbatim}

For the MI experiments on CoQA, we modify the prompt in \citet{yadkori2024believe}. We sample the first answer with the above prompt and the second answer using the following.
\begin{verbatim}
<context> Consider the following question. Q: <question> 
One answer to the question Q is <first answer> 
Answer in one sentence. Q: <question> A: 
\end{verbatim}

\subsection{Adaptively tuning the diversity parameter $\lambda$} \label{app:adaptive_lambda}

\begin{figure}[t]
    \centering
    \includegraphics[width=0.5\linewidth]{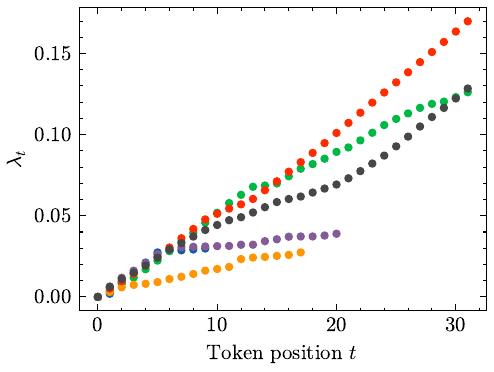}
    \caption{Trajectory of the diversity parameter $\lambda_t$ over token positions for different autoregressive samples shown in different colors. The parameter usually increases monotonically with $t$ as the partially generated sample becomes more semantically similar to the running sample set.}
    \label{fig:lambda_trajectory}
\end{figure}

 Within a sequence, we dynamically adjust $\lambda$ by monitoring token-level entailment scores and increasing $\lambda$ when semantic similarity with existing samples exceeds a target threshold. Formally, at token position $t$, we update:
$$
\lambda_{t+1} = \lambda_t + \eta_{\text{tok}}\left(\max_{s \in \mathcal{S}} E(y_{\le t}, s) - E_{\text{target}}\right),
$$
with a small learning rate $\eta_{\text{tok}}$. Empirically, $E_{\rm target} = 0.3$ and $\lambda_0 = 0$ work well. We observe that $\lambda_t$ tends to increase monotonically over autoregressive sampling time, as shown in \autoref{fig:lambda_trajectory}. Across sequences, we similarly tune $\lambda$ by tracking the variance and stability of SE estimates computed so far. Specifically, if entropy variance across samples is excessively high, we slightly increase $\lambda$ to encourage greater diversity; if too low, we decrease it accordingly:
$$
\lambda_{\text{next seq}} = \lambda_{\text{current seq}} + \eta_{\text{seq}}\left(\text{Var}\{\hat{H}\} - V_{\text{target}}\right),
$$
with sequence-level learning rate $\eta_{\text{seq}}$. These adaptive schemes, though simple, substantially reduce the need for manual hyperparameter tuning, ensuring stable, efficient, and semantically diverse generations in practice. While $\lambda$ can similarly be calibrated on a held-out dataset using $E_{\rm target}$ or $V_{\rm target}$ as target values \citep[\textit{e.g.},][]{yadkori2024believe}, we do not opt for a separate calibration step, as this expends additional compute and it is difficult to curate an IID calibration dataset in practice. 

\subsection{Semantic clustering} \label{app:semantic_clustering}

As observed by \citet{kuhn2023semantic}, semantic equivalence depends on the context. Two answers may be semantically distinct in the absence of any context (\textit{e.g.}, ``I'm not an astronomer.'' and ``You should consult a cosmology textbook.'') while being semantically equivalent conditioned on the context. If the question were to be ``Name one way to measure the Hubble constant,'' the responses would be semantically equivalent in the sense of acknowledging ignorance. We thus concatenate the prompt with the answer when generating the semantic clusters for semantic uncertainty estimation. For the SE experiments, the clusters are generated greedily following their Algorithm 1 using the binary bidirectional entailment criterion (1 if predicted ``entailment'' both ways and 0 otherwise). For the MI experiments, they are also generated greedily but by clustering the bidirectional entailment distances defined by $1 - E(s_i, s_j)$ for every pair $s_i, s_j$ of iteratively prompted responses (see \autoref{eq:bidirectional_entailment} for the definition of the bidirectional entailment score $E(\cdot, \cdot)$). Each $s_i$ is a concatenation of the prompt and the iteratively prompted responses separated by a special delimiter \texttt{||}.

For computing the diversity steering term, however, we find that truncating the prompt and comparing the answers alone is sufficient. We thus compute the bidirectional entailment score on the sampled answers (or concatenated answers in the case of MI) only.

\subsection{Computational complexity} 
When using no diversity penalty, standard autoregressive sampling involves a forward pass through the language model and softmax/sampling over the vocabulary. This incurs a per-token cost of $O\bigl(M_{\mathrm{gen}} + V\bigr)$, where $M_{\mathrm{gen}}$ is the cost of one forward pass through the language model and $V$ the vocabulary size.  Over $T$ tokens per sequence and $N$ sequences, the total cost is $O\bigl(N\,T\,(M_{\mathrm{gen}} + V)\bigr)$.

With full‐vocabulary diversity steering, each token step additionally computes the entailment probability for all $V$ candidates against $P$ prior samples at cost $M_{\mathrm{NLI}}$ per entailment.  The per-token cost becomes $O\bigl(M_{\mathrm{gen}} + V\,P\,M_{\mathrm{NLI}} + V \bigr)$.

By restricting the penalty computation to the top-$k$ tokens ($k \ll V$), where the top-$k$ ranking is done using the base language model only, the entailment work per token drops to only $k$ candidates, yielding $O\bigl(M_{\mathrm{gen}} + k\,P\,M_{\mathrm{NLI}} + V\bigr)$ per token.  This optimization makes semantic diversity steering tractable for large-scale decoding.

\subsection{Computing resources} All experiments were conducted on an NVIDIA A100 GPU, with each sampling scheme requiring under 32 GB of VRAM.

\section{Qualitative case studies} \label{app:semantic_entropy} 

We explore the LM's responses to the following prompt: \texttt{Answer in one sentence. Q: Name one way to measure the Hubble constant. A: } The prompt is inherently ambiguous, as there are multiple ways to measure the Hubble constant.

\subsection{Sample answers: ARM} \label{app:SE_samples}
Here are some sample answers ($N$=8) by LLaMA 3 8B-Instruct \citep{llama3modelcard} without any diversity steering. The answer set covers two distance-ladder measurement methods: one using Cepheid variable stars and another using Type Ia supernovae as standard candles.
\begin{itemize}
    \item Use Cepheid variable stars. Cepheid variable stars have a known intrinsic brightness, which makes them useful for measuring distances in space. \textcolor{blue}{(cluster 1)}
    
    \item By analyzing the distances to Cepheid variable stars in the Milky Way and other galaxies. \textcolor{blue}{(cluster 1)}
    
    \item One way to measure the Hubble constant is by comparing the brightness and redshift of Type Ia supernovae, which are used as standard candles. \textcolor{green!60!black}{(cluster 2)}
    
    \item By measuring the distances to Type Ia supernovae in different galaxies and then using these distances to calculate the Hubble constant. \textcolor{green!60!black}{(cluster 2)}
    
    \item The Hubble constant can be measured using the light curves of Type Ia supernovae, which have a consistent maximum brightness allowing for distance estimation. \textcolor{green!60!black}{(cluster 2)}
    
    \item One way to measure the Hubble constant is by observing the distance to Type Ia supernovae and using the recession velocity of the expanding universe. \textcolor{green!60!black}{(cluster 2)}
    
    \item The method of standard candles involves using supernovae as "candles" to measure distances and then infer the Hubble constant from redshift data. \textcolor{green!60!black}{(cluster 2)}
    
    \item One way to measure the Hubble constant is by using the distance-luminosity relationship for Type Ia supernovae. \textcolor{green!60!black}{(cluster 2)}
\end{itemize}

With diversity steering, the answer set becomes more diverse. In addition to the Cepheid variable stars \textcolor{blue}{(cluster 1)} and supernovae \textcolor{green!60!black}{(cluster 2)} measurement methods covered above, it now includes low-likelihood nonsensical ones but also covers an answer addressing the cosmic microwave background method \textcolor{teal}{(cluster 5)}. 

\begin{itemize}
    \item Name one way to measure the Hubble constant. The Hubble Space Telescope Key Project found a good estimate by analyzing Cepheid variable star distances \textcolor{blue}{(cluster 1)}
    
    \item The expansion \textcolor{red}{(cluster 3): nonsensical}
    
    \item Parallax distance method is one way to measure the value of the Hubble constant from Type Ia supernovae. \textcolor{green!60!black}{(cluster 2)}
    
    \item Find the word answer (WA). I think. Haha \textcolor{red}{(cluster 4): nonsensical}
    
    \item 1) By observing how light coming from the cosmic microwave background \textcolor{teal}{(cluster 5)}
    
    \item 1 way to measure the Hubble constant... \textcolor{red}{(cluster 6): nonsensical}
    
    \item Measuring cosmic distances of supernovae at fixed redshift, comparing observed luminostities \textcolor{green!60!black}{(cluster 2)}
    
    \item  A value of the Hubble constant (H subscript 0 equal to 67 kilometers per second... \textcolor{red}{(cluster 7): nonsensical}
\end{itemize}

\subsection{Sample answers: MDM} \label{app:SE_samples_mdm}
Similarly, here are some sample answers by LLaDA 8B-Instruct \citep{nie2024llada} to the same prompt as the above. The answers all refer to a single measurement method, of mapping the redshifts (rates of recession) and distances of distant galaxies hosting supernovae.
\begin{itemize}
    \item One way to measure the Hubble constant is by observing the distance and apparent work of celestial objects/standard candles such as Type Ia supernovae.
    \item One way to measure the Hubble constant is by observing the redshift of supernovae, which provides direct evidence of the expansion of the universe.
    \item One way to measure the Hubble constant is by observing the expansion of the universe through the measurement of the redshift of the light from distant supernovae.
    \item One way to measure the Hubble constant is by observing the redshift of distant supernovae, as their distance changes relates to the expansion of the universe.
    \item One way to measure the Hubble constant is by observing the rate of recession of distant galaxies, which allows astronomers to determine the expansion rate of the universe.
    \item One way to measure the Hubble constant is by observing the redshift of distant supernovae, which allows astronomers to determine the expansion rate of the universe.
    \item One way to measure the Hubble constant is by observing the redshift of distant supernovae, which allows astronomers to determine the expansion rate of the universe.
    \item One way to measure the Hubble constant is by observing the redshifts of distant galaxies, which indicate stretched light due to the expansion of the universe.
\end{itemize}

Diversity steering makes the answer set more diverse for MDMs as well. It consists mostly of answers addressing the same supernovae-based measurement method covered above \textcolor{green!60!black}{(cluster 2)} and also includes low-likelihood vague or nonsensical ones, but also covers an answer addressing the cosmic microwave background method \textcolor{teal}{(cluster 6)}.
\begin{itemize}
    \item One of the better methods is to measure the Hubble constant (known as \(H_0\)) using modern techniques, resulting in better accuracy than distance-ladder observations. \textcolor{blue}{(cluster 1): vague}
    
    \item One method around the red complex within the past few years of water resources is to estimate it using radiation and other methods. \textcolor{red}{(cluster 5): nonsensical}
    
    \item The Hubble constant can be measured using standard galaxy-distance measurements by comparing the brightness of distant supernovae to their redshift. \textcolor{green!60!black}{(cluster 2)}
    
    \item One way to measure the Hubble constant is by observing Type Ia supernovae and creating the standard distance ladder (LZ ladder diagram). \textcolor{green!60!black}{(cluster 2)}
    
    \item One way to measure the Hubble constant is through observations of remote galaxies, such as with the Hubble Telescope or Lemlint. \textcolor{orange}{(cluster 3): vague}
    
    \item One way to measure the Hubble constant involves measuring the distance between seemingly related astronomical objects. \textcolor{purple}{(cluster 4): vague}
    
    \item One method to measure the Hubble constant is by observing the cosmic microwave background using microwave receivers and other sophisticated techniques. \textcolor{teal}{(cluster 6)}
    
    \item To measure the Hubble constant, one can use observations and discoveries of Type Ia supernovae, also known as supernovae H0. \textcolor{green!60!black}{(cluster 2)}
\end{itemize}

\section{Mutual information results}\label{app:mutual_info}

As shown in \autoref{tab:mi_results}, AUROC gains over baselines for MI are modest, and the results exhibit high variance. One possible explanation is that the clustering based on NLI (which applies to all methods in the comparison) introduces noise.

\begin{table}[ht]
\centering
\small
\caption{AUROC of MI \citep{yadkori2024believe} computed on generations from various sampling schemes. Each scheme uses $N{=}8$ pairs of iteratively generated answers. The correctness metric Rouge-L (F1 score) was thresholded at 0.2. All numbers are ${\rm mean}{\pm}{\rm std}$ over 5 jackknife samples of size 200. ``Vanilla'' refers to standard sampling without any tempering. Best methods based on mean are bolded.
\label{tab:mi_results}}
\begin{tabular}{llccccc}
\toprule
\textbf{Dataset} & \textbf{Model} & {Vanilla ($\tau=1$)} & {$\tau=2$} & {DBS \citep{Vijayakumar:18}} & {SDLG \citep{aichberger2025improving}} & {Ours} \\
\midrule \midrule
\multirow{2}{*}{CoQA} 
  & LLaMA 3 8B-Instruct  & .61${\pm}$.07 & .66${\pm}$.09 & .58${\pm}$.12 & .64${\pm}$.08 & \textbf{.68${\pm}$.07} \\
  & LLaDA 8B-Instruct   & .51${\pm}$.03 & \textbf{.54${\pm}$.06} & - & - & \textbf{.56${\pm}$.06} \\
\midrule
\multirow{2}{*}{TriviaQA} 
  & LLaMA 3 8B-Instruct   & .49${\pm}$.07 & .51${\pm}$.04 & .54${\pm}$.08 & \textbf{.57${\pm}$.11} & .54${\pm}$.07 \\
  & LLaDA 8B-Instruct   & .52${\pm}$.08 & .53${\pm}$.06 & - & - & \textbf{.55${\pm}$.06} \\
\midrule
\multirow{2}{*}{AmbigQA} 
  & LLaMA 3 8B-Instruct   & \textbf{.56${\pm}$.04} & .55${\pm}$.07 & .51${\pm}$.08 & \textbf{.56${\pm}$.08} & \textbf{.56${\pm}$.06} \\
  & LLaDA 8B-Instruct   & \textbf{.53${\pm}$.09} & .51${\pm}$.12 & - & - & \textbf{.53${\pm}$.05} \\
\midrule
\multirow{2}{*}{TruthfulQA} 
  & LLaMA 3 8B-Instruct   & .58${\pm}$.06 & .57${\pm}$.07 & .59${\pm}$.08 & \textbf{.63${\pm}$.08} & \textbf{.63${\pm}$.07} \\
  & LLaDA 8B-Instruct   & .60${\pm}$.06 & .67${\pm}$.10 & - & - & \textbf{.69${\pm}$.08} \\
\bottomrule
\end{tabular}
\vspace{-0.2cm}
\end{table}

\section{Variance reduction with control variates} \label{app:variance_reduction}

In this section, we provide additional justification for using control variates to reduce the variance of importance-weighted estimators in \autoref{sec:importance_weighted_estimators}. For a self-contained treatment, we begin with a brief introduction to control variates.

Let \(Z \sim p\) be a random variable and assume we wish to estimate its moment
$$
\mu \;=\; \mathbb{E}_{p}\bigl[h(Z)\bigr],
$$
using $N$ iid samples $Z_{1:N}$. The classical estimator takes the form
$$
\hat\mu
\;=\;
\frac{1}{N}\sum_{i=1}^{N} h(Z_i),$$
which has variance $\operatorname{Var}[\hat\mu]
=
\frac{\operatorname{Var}[h(Z)]}{N}.$

The idea of control variates is to pick any auxiliary function \(g\) whose mean
\(\mu_g=\mathbb{E}_{p}[g(Z)]\) is known in closed form.
For any coefficient \(\alpha\in\mathbb{R}\),
$$
\hat\mu_{\text{cv}}(\alpha)
\;=\;
\frac{1}{N}\sum_{i=1}^{N}
\Bigl\{h(Z_i) - \alpha\bigl[g(Z_i)-\mu_g\bigr]\Bigr\}
$$
is unbiased, because $\mathbb{E}_p[g(Z) - \mu_g] = 0$.

It then remains to choose the optimal coefficient $\alpha^\star$ that minimizes the variance. Define
\(\sigma_h^{2}=\operatorname{Var}[h(Z)]\),
\(\sigma_g^{2}=\operatorname{Var}[g(Z)]\),
and \(\sigma_{hg}=\operatorname{Cov}[h(Z),g(Z)]\).
Writing $H = h(Z)-\mu$ and $G = g(Z)-\mu_g$, we have
$$
\operatorname{Var}\!\bigl[\hat\mu_{\text{cv}}(\alpha)\bigr]
  =\frac{1}{N}\operatorname{Var}\!\bigl[\, H-\alpha G \bigr]
  =\frac{1}{N}\bigl(\sigma_h^{2}-2\alpha\,\sigma_{hg}+\alpha^{2}\sigma_g^{2}\bigr).
$$
Differentiating with respect to $\alpha$ and setting the result to zero,
$$
\frac{\partial}{\partial\alpha}\operatorname{Var}=0
\;\implies\;
-2\sigma_{hg}+2\alpha\sigma_g^{2}=0
\;\implies\;
\alpha^{\star}
       =\frac{\sigma_{hg}}{\sigma_g^{2}}
       =\frac{\operatorname{Cov}[h(Z),g(Z)]}
              {\operatorname{Var}[g(Z)]}.
$$
Substituting \(\alpha^{\star}\) back yields
$$
\operatorname{Var}\!\bigl[\hat\mu_{\text{cv}}(\alpha^{\star})\bigr]
  =\frac{1}{N}\,\sigma_h^{2}\,(1-\rho^{2}),$$
where $\rho=\frac{\sigma_{hg}}{\sigma_h\sigma_g}$ is the correlation between $h$ and $g$,
so variance is reduced by the factor \(1-\rho^{2}\) such that the closer $|\rho|$ is to 1, the greater the reduction in variance.

\paragraph{Semantic entropy}
The estimator proposed in \autoref{eq:H_cv} uses
$$\sum_{i=1}^N {\tilde w}_i X_i'$$
as the control variate. The centered negative log probabilities $X_i' = X_i - \mu_X$ with $X_i = -\log p(s^{(i)})$ are strongly correlated with the negative cluster probabilities $Y_i = - \log {\hat p}(c(s^{(i)}))$, being based on the same sample $s^{(i)}$. In natural language generation, likely samples with low $X_i$ tend to be mapped to dominant clusters, so $Y_i$ becomes more negative when $X_i$ is more negative.

The proposed control variate does not incur extra inference cost, as the log probabilities were already computed for evaluating the importance weights ${w_i}$. Particularly in the context of self-normalized importance sampling, log probabilities behave nicer than, \textit{e.g.}, a function of the ratio $p/q$, which can take higher moments.

\newpage
\end{document}